\def\year{2020}\relax
\documentclass[letterpaper]{article} 
\usepackage{aaai20}  
\usepackage{times}  
\usepackage{helvet} 
\usepackage{courier}  
\usepackage[hyphens]{url}  
\usepackage{graphicx} 
\usepackage{amsmath}
\usepackage{multirow}
\usepackage{array}
\usepackage{amsfonts,amssymb}
\usepackage{pdfpages}

\title{CatGAN: Category-aware Generative Adversarial Networks with
Hierarchical Evolutionary Learning for Category Text Generation}
\author{Zhiyue Liu, Jiahai Wang,\thanks{Corresponding author} Zhiwei Liang\\
School of Data and Computer Science, Sun Yat-sen University, Guangzhou, China\\
\{liuzhy93, liangzhw25\}@mail2.sysu.edu.cn,
wangjiah@mail.sysu.edu.cn}

\begin{document}
\maketitle
\begin{abstract}
Generating multiple categories of texts is a challenging task and draws more and more attention. Since generative adversarial nets (GANs) have shown competitive results on general text generation, they are extended for category text generation in some previous works.
However, the complicated model structures and learning strategies limit their performance and exacerbate the training instability.
This paper proposes a category-aware GAN (CatGAN) which consists of an efficient category-aware model for category text generation and a hierarchical evolutionary learning algorithm for training our model.
The category-aware model directly measures the gap between real samples and generated samples on each category, then reducing this gap will guide the model to generate high-quality category samples.
The Gumbel-Softmax relaxation further frees our model from complicated learning strategies for updating CatGAN on discrete data.
Moreover, only focusing on the sample quality normally leads the mode collapse problem, thus a hierarchical evolutionary learning algorithm is introduced to stabilize the training procedure and obtain the trade-off between quality and diversity while training CatGAN.
Experimental results demonstrate that CatGAN outperforms most of the existing state-of-the-art methods.

\end{abstract}

\section{Introduction}
Nowadays, category text generation has received more and more attention. Generating coherent and meaningful text with different categories will bring great benefits to many natural language processing applications, such as sentiment analysis~\cite{li2018generative} and dialogue generation~\cite{li2017adversarial}.
Recently, generative adversarial net (GAN)~\cite{goodfellow2014generative}, which adopts the discriminator to guide the generator, is combined with the reinforcement learning (RL) algorithms~\cite{williams1992simple} to generate discrete text data for general text generation, and some competitive results have been reported in the previous works~\cite{yu2017seqgan,guo2018long,caccia2018language}.
Compared with general text generation which only focuses on obtaining high-quality text, category text generation aims at automatically generating a variety of controllable category text to fit the task-specific applications.
However, the category information of sentences can not be easily controlled, and it is also difficult to design an appropriate training objective for different categories. Thus, category text generation is a more challenging task.
There are a few works~\cite{wang2018sentigan,li2018generative} which try to extend the general text generation models for category text generation. They mostly employ a long short-term memory (LSTM)~\cite{hochreiter1997long} as the generator and combine the auxiliary components (e.g., classifiers) with the RL algorithms on GANs to generate category text. The auxiliary components can help the model to focus on the category information.

The existing category text generation models have shown some positive results, but RL algorithms and auxiliary components complicate the learning strategy and the model, respectively, which may exacerbate some fundamental problems of GANs, including training instability and mode collapse. Firstly, most of the existing models~\cite{wang2018sentigan,li2018generative} heavily rely on RL algorithms, and some strategies, such as Monte Carlo search, are adopted to guide the discriminator for providing reward signals. These complicated strategies further increase the training difficulty of GANs. The auxiliary components may carry more burden to the adversarial training, which also makes the training procedure more unstable. Secondly, the mode collapse problem is serious in the existing models. Because the LSTM based generator~\cite{li2018generative} may lack enough expressive power, and category text generation, as a sequential decision process, also easily leads the generator to focus on some limited samples in the target distribution. For generating diversified samples, the temperature variable~\cite{caccia2018language,guo2018long} is employed to make GANs focus on either the quality or the diversity, but an improvement of diversity always leads to significant degradation of quality.

In this paper, a new category text generation framework, category-aware GAN (CatGAN), is proposed to deal with the above problems. CatGAN provides a category-aware model for category text generation and a hierarchical evolutionary learning algorithm for training the model and obtaining the balance between the sample quality and diversity.
Firstly, a novel category-aware model is proposed, which includes the category-wise relativistic objective to estimate the gap between the specific category generated samples and the corresponding real samples. The generator wants to make the generated samples as realistic as the real samples, while the discriminator is eager to enlarging this gap. The relativistic relation can guide our model to update more easily than strict ground-truth labels. A relational memory core (RMC)~\cite{santoro2018relational} based generator, which promises a larger memory capacity and a better ability for catching the long-term dependencies, is adopted to replace LSTM. Further, instead of RL algorithms, CatGAN employs the Gumbel-Softmax relaxation~\cite{jang2016categorical,maddison2016concrete} to generate the continuous approximation of the discrete generated samples. The continuous data allow the generator and the discriminator to be optimized directly during the adversarial procedure. Without any auxiliary components, the architecture of CatGAN is as concise as the classical GAN framework and only consists of one generator and one discriminator.
Secondly, a hierarchical evolutionary learning algorithm is developed to train the category-aware model.
The adversarial training can be seen as an evolutionary problem, and the discriminator provides the environment for a population of generators to evolve.
For adapting the category text generation task, the evolution procedure is designed with two stages.
In the first temperature-oriented stage, the temperature is subtly controlled to maintain the category text quality during the improvement of diversity.
In the second objective-oriented stage, various training objectives are adopted to narrow the distances between the generated data and the real data from different perspectives on each category. Only the well-performing generator is preserved, and the generated samples will retain diversified and high-quality.
Finally, although the evaluation metric of quality has been designed well~\cite{guo2018long}, the evaluation metric of diversity is not explored well.
This paper proposes a new evaluation metric of diversity based on the repeatability of the generated samples.

In summary, our contributions are as follows:
\begin{quote}
\begin{itemize}
\item A category-aware model is proposed for generating category text, which accurately takes the gap between real samples and generated samples on each category as an efficient learning signal.

\item A hierarchical evolutionary learning algorithm is designed to train the category-aware model, and it specializes in text generation for making the generated samples more diversified and high-quality. 

\item An effective metric is presented to evaluate the sample diversity. Experimental results on synthetic and real data demonstrate that our model achieves a new state-of-the-art performance on both category text generation and general text generation.
\end{itemize}
\end{quote}

\section{Related Work}
Traditional recurrent neural network (RNN) based text generation models~\cite{graves2013generating} always suffer from the exposure bias problem~\cite{huszar2015not,bengio2015scheduled}.
Different from these RNN based models which are trained by maximum likelihood estimation (MLE), GAN introduces a
minimax game between the generator and the discriminator. However, GAN is designed to output differentiable data, which has a conflict with the discrete text generation.

The same RL algorithm is adopted in SeqGAN~\cite{yu2017seqgan} and LeakGAN~\cite{guo2018long} to solve the above problem, and the discriminator can guide the generator by the reward signal. However, LeakGAN shows that the reward signal is not sufficiently informative. MaskGAN~\cite{fedus2018maskgan} adopts the actor-critic algorithm for filling in missing text conditioned on the surrounding context. RankGAN~\cite{lin2017adversarial} replaces the original binary classifier with a ranking model as the discriminator. Approximating methods are another way to handle the non-differentiable problem of discrete data. TextGAN~\cite{zhang2017adversarial} and FM-GAN~\cite{chen2018adversarial} apply an annealed softmax to approximate the argmax operation. Gu et al.~\shortcite{gu2018neural} and RelGAN~\cite{nie2018relgan} adopt the Gumbel-Softmax relaxation to approximate the categorical distribution, and this relaxation method helps to train GANs and improve the generation quality.

The above methods focus on general text generation, and category text generation is drawing more attention. CSGAN~\cite{li2018generative} proposes a descriptor which consists of a discriminator and an auxiliary classifier, where the classifier distinguishes the sentence category to guide the generator. The adversarial procedure of CSGAN is similar to SeqGAN, and the less-informative reward signal limits the model performance.
SentiGAN~\cite{wang2018sentigan} contains multiple generators, and each generator aims at generating the samples of a specific sentiment label. However, as the category number grows up, the multiple generators will significantly raise the number of trainable parameters, which may reduce the efficiency and amplify the training instability. Experiments will show that the proposed CatGAN is more effective than the previous methods using auxiliary components.

Recently, the evolutionary learning algorithm is firstly introduced to optimize the adversarial model for image generation~\cite{wang2019evolutionary}. For generating better category text, both the quality and diversity should be focused. CatGAN makes the first attempt to solve category text generation with the evolutionary learning algorithm. Our hierarchical evolutionary learning algorithm is designed with two stages, the temperature-oriented stage and the objective-oriented stage, to explore the possible solutions of the generator for improving the performance on the sample quality and diversity.

\begin{figure*}[t]
\centering
\includegraphics[width=1.96\columnwidth]{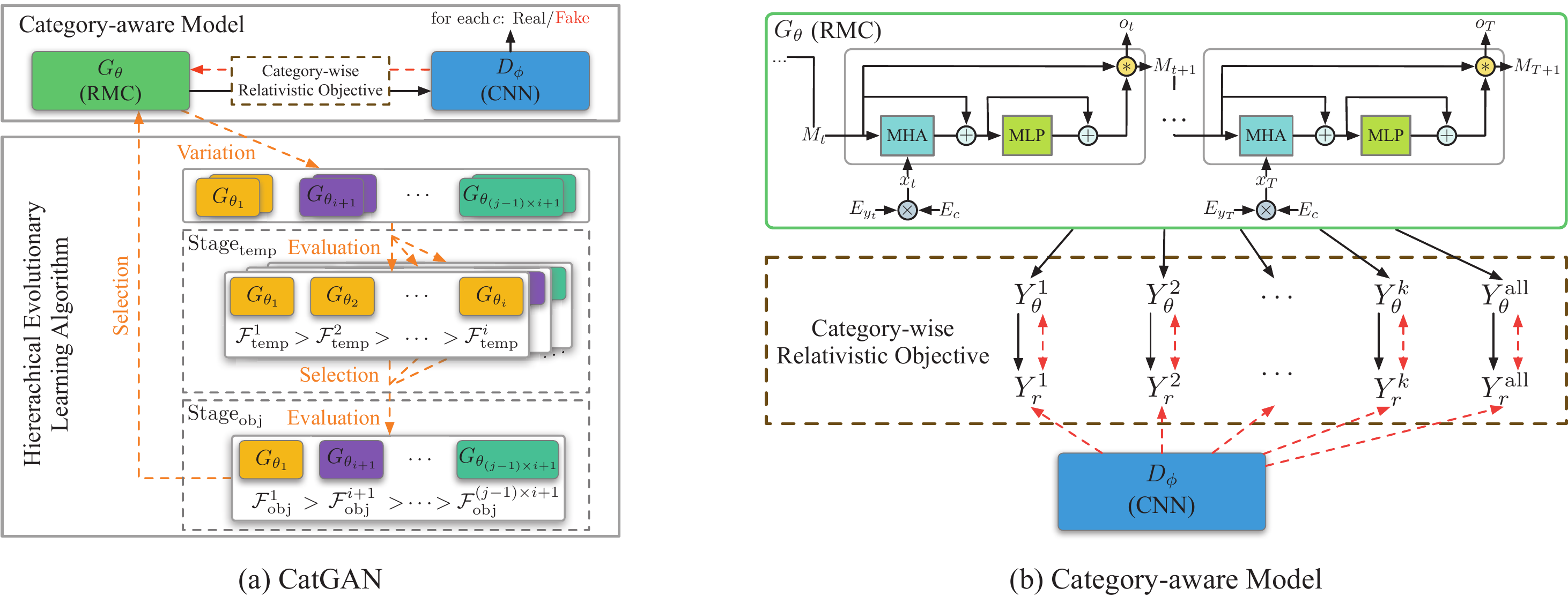}
    \caption{(a) The overall framework of CatGAN. A population of generators $\{G_\theta\}$ evolves in a dynamic environment denoted by the discriminator $D_\phi$. In each round of evolution, the individuals $\{G_{\theta_1},G_{\theta_2},\cdots\}$ after the variation process undergo two stages of evaluation and selection to hierarchically form a new population, where $i = |\mathbb{T}|$ and $j = |\mathbb{O}|$. The individuals, mutated according to the same training objective under different temperatures, are denoted with the same color. (b) Category-aware model. The red dotted line represents the training process of the discriminator, while the black full line represents the training process of the generator. The MHA denotes the multi-head dot product attention.}
    \label{fig:CatGAN}
\end{figure*}

\section{Methodology}
The category text generation task is denoted as follows.
Given a dataset with $k$ categories, supposing we want to generate a sentence with the specific category $c$, then a $\theta$-parameterized generator $G_{\theta}$ is trained to generate the sentence $Y_{\theta}^{c} = (y_1, \ldots, y_t, \ldots, y_T)$, $y_t \in \mathcal{V}$, where $\mathcal{V}$ is the vocabulary of candidate tokens.
In order to guide the generator $G_{\theta}$ effectively, a $\phi$-parameterized discriminator $D_{\phi}$ also need to be trained to provide a learning signal $D(Y_{\theta}^{c})$ for $G_{\theta}$ to update when the whole sentence $Y_{\theta}^{c}$ has been generated.

\subsection{Overall Framework}
The overall framework of CatGAN is shown in Fig.~\ref{fig:CatGAN} (a).
CatGAN consists of two core parts, the category-aware model and the hierarchical evolutionary learning algorithm.

With the help of the category-wise relativistic objective, the proposed category-aware model employs a RMC based generator to generate texts with a specific category $c$ to fool the discriminator, while the discriminator is trained to discriminate between the real samples and the generated samples for each category. In our model, the Gumbel-Softmax relaxation enables the gradients to pass back to the generator from the discriminator directly. For training the model and boosting the performance, this paper proposes the hierarchical evolutionary learning algorithm, which evolves a population of generators $\{G_{\theta}\}$ via combining various mutation strategies in a given environment $D_{\phi}$.
At the end of the evolutionary learning algorithm, the best-performing generator is preserved to generate the realistic sentences with the given category.

\subsection{Category-aware Model}
The category-aware model is shown in Fig.~\ref{fig:CatGAN} (b), which is guided by a novel category-wise relativistic objective to generate category samples. It includes a generator $G_{\theta}$ and a basic CNN based discriminator $D_{\phi}$~\cite{kim2014convolutional}.

\subsubsection{Category-wise Relativistic Objective.}

In the standard GAN~\cite{goodfellow2014generative}, the discriminator is trained on the ground-truth labels to predict the probability that the input data are real. By this training method, the discriminator cannot provide an informative signal to update the generator~\cite{jolicoeur2018relativistic}. Thus, this paper proposes a novel training objective based on the relativistic relation between real data and generated data on each category.

Formally,  $Y_{r}^{c}$ and $Y_{\theta}^{c}$ denote the real data sampled from the real data distribution $P_{r}^{c}$ and the generated data sampled from the generated data distribution $P_{\theta}^{c}$ on the category $c$, respectively.
The category-wise relativistic objective contains the summed category loss and the enhanced real-fake loss. For the discriminator objective, it is defined as follows:
\begin{equation}
\resizebox{0.80\columnwidth}{!}{$
l_{D_{\phi}}^{\text{CatRa}} = \sum_{c=1}^{k} \mathcal{L}^{\text{Ra}}(Y_{r}^c, Y_{\theta}^c) + \mathcal{L}^{\text{Ra}}(Y_{r}^{\text{all}}, Y_{\theta}^{\text{all}}),
$}
\label{eq:dis_loss}
\end{equation}
where $Y_{r}^{\text{all}}$ and $Y_{\theta}^{\text{all}}$ are sampled from the real data distribution $P_{r}^{\text{all}}$ and the generated data distribution $P_{\theta}^{\text{all}}$ on all categories, respectively.
On the right-hand side, the first term measures the distance between the real data and the generated data on each category, while the second term measures the overall distance on all categories.
Similar to the form of RaGAN~\cite{jolicoeur2018relativistic}, $\mathcal{L}^{\text{Ra}}(\cdot)$ is defined by:
\begin{equation}
\resizebox{.95\columnwidth}{!}{$
\begin{aligned}
& \mathcal{L}^{\text{Ra}} (Y_r, Y_{\theta}) = \\
& -\mathbb{E}_{Y_{r} \sim P_{r}}\left[\log \left( \bar{D}_{\phi}(Y_{r}) \right) \right] - \mathbb{E}_{{Y}_{\theta} \sim P_{\theta}} \left[\log \left(1- \bar{D}_{\phi}(Y_{\theta}) \right) \right],
\end{aligned}
$}
\end{equation}
where the relativistic relation is measured by:
\begin{equation}
\resizebox{.95\columnwidth}{!}{$
\begin{aligned} 
\bar{D}_{\phi}(Y)= 
\begin{cases}
\text{sigmoid}(D_{\phi}(Y)-\mathbb{E}_{Y_{\theta} \sim P_{\theta}} [D_{\phi}(Y_{\theta})])& \text{if } Y \text{ is real}\\
\text{sigmoid}(D_{\phi}(Y)-\mathbb{E}_{Y_{r} \sim P_{r}} [D_{\phi}(Y_{r})]) & \text{otherwise}.
\end{cases}
\end{aligned}
$}
\label{eq:dis_sup}
\end{equation}

Intuitively, the relativistic relation shows the gap between the probabilities of being real on the real samples and that on the generated samples. For each category, the generator wants to reduce this gap for making generated samples as realistic as real samples, while the discriminator wants to increase the probability that real samples are more realistic than generated samples. Generally, the generator objective can be set to $l_{G_{\theta}}^{\text{CatRa}}= -l_{D_{\phi}}^{\text{CatRa}}$.
Compared with the standard GAN objective, the category-wise relativistic objective can efficiently train our model for category text generation.

\subsubsection{Relational Memory Core based Generator.}
Since the LSTM based generator may lack enough expressive power for text generation, relational memory core (RMC) is employed as the generator $G_{\theta}$.
The basic concept of RMC is to consider a fixed set of memory slots (e.g., memory matrix) and allow self-attention mechanism~\cite{vaswani2017attention} to interact in these memories.
The increased capacity of memory boosts the expressive power and the ability to capture the category information. Given a new vocabulary observation $y_t$ at time $t$, it is represented by the embedded token $E_{y_t}$, and the embedded category $E_c$ is built to control the category information. Then, the input vector $x_t$ of the generator is obtained by a linear transformation $W_x$ on the concatenation of $E_{y_t}$ and $E_c$:
\begin{equation}
x_t = [E_{y_t};E_c]W_x,
\end{equation}
where $[;]$ denotes the row-wise concatenation.

Considering a memory matrix $M_t$, Fig.~\ref{fig:CatGAN} (b) shows how $M_{t+1}$ is updated from $M_t$ by incorporating $x_t$ at time $t$.
As implied by the name of the multi-head dot product attention (MHA), a $H$-heads RMC contains $H$ groups of linear transformation weights for query $M_{t}W_{q}$, key $[M_{t};x_t]W_{k}$ and value $[M_{t};x_t]W_{v}$. Then, $\tilde{M}_{t+1}$ can be interpreted as a proposed update to $M_{t}$ as follows:
\begin{equation} \label{eq:rmc_update}
\tilde{M}_{t+1} = \sigma \left( \frac{M_{t}W_{q}([M_{t};x_{t}]W_{k})^{T}}{\sqrt{d_{k}}} \right)[M_{t};x_{t}]W_{v},
\end{equation}
where $\sigma(\cdot)$ denotes the row-wise softmax function, and
$d_{k}$ is the column dimension of $[M_{t};x_t]W_{k}$.
Thus, the next memory $M_{t+1}$ and the generator output $o_t$ are obtained by:
\begin{equation} \label{eq:next_memory}
M_{t+1} = \psi_{1}(\tilde{M}_{t+1}, M_{t}), o_{t} = \psi_{2} (\tilde{M}_{t+1}, M_{t}),
\end{equation}
respectively, where the two parameterized functions $\psi_{1}(\cdot)$ and $\psi_{2}(\cdot)$ both represent the interactions between $\tilde{M}_{t+1}$ and $M_{t}$ by leveraging residual connections, multi-layer perception (MLP) and gated operations. 

Directly sampling $y_{t+1}$ from the multinomial distribution $\sigma (o_t)$ will cause the non-differentiability problem~\cite{yu2017seqgan}, thus the Gumbel-Softmax relaxation is employed with the generator to approximate the samples.
The Gumbel-Max trick~\cite{maddison2016concrete} and the softmax function $\sigma(\cdot)$ are used to sample discrete sentences and approximate the argmax function, respectively.
The Gumbel-Max trick samples the discrete token $y_{t+1}$ at time $t$ by:
\begin{equation}
y_{t+1} = \mathop{\textrm{argmax}}_{1 \leq d \leq |\mathcal{V}|} (o_{t}^{(d)} + g_{t}^{(d)}),
\end{equation}
where $o_{t}^{(d)}$ is the value of the $d$-th dimension of $o_{t}$, and $g_{t}^{(d)}$ is sampled from the Gumbel distribution, where  $g_{t}^{(d)} = -\log(-\log U_{t}^{(d)})$ with $U_{t}^{(d)} \sim \textrm{Uniform}(0,1)$.
The differentiable approximation of argmax is obtained by:
\begin{equation}
\hat{y}_{t+1} = \sigma (\tau (o_t + g_t)),
\end{equation}
where $\tau$ is the temperature variable.
Since the softmax-like token $\hat{y}_{t+1}$ is differentiable with respect to $o_t$, it is used as the input of the discriminator instead of the discrete token $y_{t+1}$.
$\tau$ can adjust bias and variance while approximating $y_{t+1}$.
Larger $\tau$ brings lower bias but higher variance~\cite{tucker2017rebar}, allowing the generator to obtain higher diversity but poorer quality samples~\cite{caccia2018language}.

\subsection{Hierarchical Evolutionary Learning Algorithm}
Unlike previous text generation methods, which adopt the fixed temperature strategy and one adversarial objective to train a generator and a discriminator, our hierarchical evolutionary learning algorithm evolves a population of generators, with various temperatures and objectives, to play the adversarial game with the discriminator.

\subsubsection{Variation.}
During the variation procedure, the individuals $\{G_{\theta_1},G_{\theta_2},\cdots\}$ are mutated from the parents $\{G_\theta\}$ via asexual reproduction based on the combination of two kinds of strategies, the temperature mutation strategy (TMS) and the objective mutation strategy (OMS).
TMS is to maintain the high sample quality when the diversity improves (i.e., $\tau$ increases).
To explore the possible solutions of the generator in the parameter space, OMS further stabilizes the model training process via leveraging various training objectives.

Previous works adopt the monotone increasing function $f_{\tau_\text{tar}}(n)$ to boost $\tau$ over training iterations, where $\tau_\text{tar}$ is the target temperature, and $n \in [1, N]$ denotes the current iteration of the maximum iterations $N$. $\tau$ obtains a subtle increment after each iteration.
Although the monotone increasing $\tau$ brings diversity, it leads to quality degradation.
In one iteration, the subtle change of $\tau$ only affects the tightness of the relaxation for one batch of training samples which cannot fully represent all training samples.
Thus, various subtle changes have the potential to improve the sample quality.
With overall increasing $\tau$, TMS aims at finding the optimal temperature change direction according to the quality in each iteration. The comparison between the evolutionary temperature and the monotone increasing temperature is given in the \mbox{\emph{Appendix C}}.
Formally, $\mathbb{T}$ denotes the TMS set that includes the temperatures with various change directions:
\begin{equation}
\mathbb{T} = \{f_{\tau_\text{tar}}(n-1), f_{\tau_\text{tar}}(n),f_{\tau_\text{tar}}(n+1)\}.
\end{equation}

Besides, $\mathbb{O}$ denotes the OMS set which contains several relativistic training objectives for the generator as follows:
\begin{equation}
\mathbb{O} = \{l_{G_{\theta}}^{\text{CatRS}},l_{G_{\theta}}^{\text{CatRa}}\},
\end{equation}
where $l_{G_{\theta}}^{\text{CatRS}}$ is another way to measure the the relativistic relation for all categories, similar to the form of $l_{G_{\theta}}^{\text{CatRa}}$:
\begin{equation}
\resizebox{.95\columnwidth}{!}{$
\begin{aligned}
& l_{G_{\theta}}^{\text{CatRS}} = 
- \sum_{c=1}^{k} \mathbb{E}_{
\begin{subarray}{l}
Y_{r}^{c} \sim P_{r}^{c} \\
Y_{\theta}^{c} \sim P_{\theta}^{c}
\end{subarray}
} [\text{log}(\text{sigmoid}(D_{\phi}(Y_{\theta}^{c}) - D_{\phi}(Y_{r}^{c})))] \\
& -\mathbb{E}_{
\begin{subarray}{l}
Y_{r}^{\text{all}} \sim P_{r}^{\text{all}} \\
Y_{\theta}^{\text{all}} \sim P_{\theta}^{\text{all}}
\end{subarray}
} [\text{log}(\text{sigmoid}(D_{\phi}(Y_{\theta}^{\text{all}}) - D_{\phi}(Y_{r}^{\text{all}})))].
\end{aligned}
$}
\end{equation}

It is worth noting that adding more temperatures and objectives into $\mathbb{T}$ and $\mathbb{O}$ is feasible. The Cartesian product of $\mathbb{T}$ and $\mathbb{O}$ constitutes all mutation directions $\mathbb{M} = \mathbb{T} \times \mathbb{O}$ on each round of evolution.
Each individual is mutated by one mutation direction. That is, the generator is updated by a specific training objective under a certain temperature.

\subsubsection{Hierarchical Evaluation.}

Since the goals of TMS and OMS are different, two stages, including the temperature-oriented stage $\textrm{Stage}_\textrm{temp}$ and the objective-oriented stage $\textrm{Stage}_\textrm{obj}$, are designed. Both the evaluation procedure and the selection procedure are divided into the above two stages, where $\textrm{Stage}_\textrm{temp}$ can preliminarily filter the individuals for $\textrm{Stage}_\textrm{obj}$. In $\textrm{Stage}_\textrm{temp}$, the individuals with different temperatures in $\mathbb{T}$ can only be compared under the same objective. For each objective in $\mathbb{O}$, the individual with the optimal temperature is preserved for the further selection. Then, $\textrm{Stage}_\textrm{obj}$ selects the best individual considering overall performance.
Thus, the proposed learning algorithm including two stages, $\textrm{Stage}_\textrm{temp}$ and $\textrm{Stage}_\textrm{obj}$, is considered as the hierarchical evolutionary learning algorithm. Two properties, the sample diversity and quality, are mainly considered to measure the performance of each individual in the whole hierarchical evolutionary learning algorithm.

For evaluating the diversity, a new metric named $\textrm{NLL}_{\textrm{div}}$ is proposed. $\textrm{NLL}_{\textrm{div}}$ calculates the negative log-likelihood of generated samples on the generator by:
\begin{equation} \label{eq:nll_self}
    \textrm{NLL}_\textrm{div}=-\mathbb{E}_{{Y_{\theta}}\sim{P_\theta}}[\log{P_\theta}(y_{1},\cdots,y_{T})],
\end{equation}
where $P_\theta$ is the generated sample distribution.
$\textrm{NLL}_{\textrm{div}}$ can captures the repeatability of the generated samples, which will better reflect the mode collapse issue.
When the generator can only learn some limited patterns from the real data or assign all its probability mass to a small region, the value of $\textrm{NLL}_{\textrm{div}}$ will become extremely low.

For evaluating the quality, $\bar{D}_{\phi}(Y_\theta)$ in Eq.~\ref{eq:dis_sup} can accurately measure the gap between the generated samples and the real samples. The higher $\bar{D}_{\phi}(Y_\theta)$, the better quality sentences that the generator can generate. Therefore, the evaluation scores in $\textrm{Stage}_\textrm{temp}$ and $\textrm{Stage}_\textrm{obj}$ are respectively defined as:
\begin{equation}
\mathcal{F}_{\textrm{temp}} = \mathbb{E}_{{Y_{\theta}}\sim{P_\theta}}[\bar{D}_{\phi}(Y_\theta)],
\label{eq:f_temp}
\end{equation}
\begin{equation}
\mathcal{F}_{\textrm{obj}} = \mathbb{E}_{{Y_{\theta}}\sim{P_\theta}}[\bar{D}_{\phi}(Y_\theta)] + \lambda \textrm{NLL}_{\textrm{div}},
\label{eq:f_obj}
\end{equation}
where $\lambda$ can be tuned to balance the quality and diversity. $\mathcal{F}_{\textrm{temp}}$ aims at maintaining the quality when $\tau$ increases in $\textrm{Stage}_\textrm{temp}$, and $\mathcal{F}_{\textrm{obj}}$ wants to stabilize the training process and further balance the sample quality and diversity. Then, we expect to maximize $\mathcal{F}_{\textrm{temp}}$ and $\mathcal{F}_{\textrm{obj}}$ hierarchically.

\subsubsection{Hierarchical Selection.}
The evaluation procedure of each stage corresponds to a selection process, which selects the individuals with larger evaluation scores. Firstly, according to each objective in $\mathbb{O}$, the individual which has the largest $\mathcal{F}_{\textrm{temp}}$ is preserved with a selected direction in $\mathbb{T}$. Secondly, the surviving individuals are further filtered based on $\mathcal{F}_{\textrm{obj}}$ to obtain the best-performing generators as the new parents, which will participate in future adversarial training. Following the principle of ``survival of the fittest'', the optimal temperature and training objective are selected for the generator, allowing the whole model is trained as expected.

\section{Experimentation}
\subsection{Experimental Setting}
\subsubsection{Evaluation Metrics.}
Some evaluation metrics have been widely used to measure the performance of text generation models from various aspects~\cite{semeniuta2018accurate}. Generally, the negative log-likelihood $\textrm{NLL}_\textrm{oracle}$~\cite{yu2017seqgan} is used to measure the quality on synthetic data. Two evaluation metrics, $\textrm{NLL}_\textrm{div}$ and $\textrm{NLL}_\textrm{gen}$, are adopted to measure the diversity, where $\textrm{NLL}_\textrm{div}$ is defined by Eq. \ref{eq:nll_self}, and $\textrm{NLL}_\textrm{gen}$~\cite{zhu2018texygen} is the reversed direction of $\textrm{NLL}_\textrm{oracle}$. $\textrm{NLL}_\textrm{oracle}$ and $\textrm{NLL}_\textrm{gen}$ are defined as follows:
\begin{align}
  \textrm{NLL}_\textrm{oracle}=-\mathbb{E}_{{Y_{\theta}}\sim{P_\theta}}[\log{P_r}(y_{1},\cdots,y_{T})], \\
  \textrm{NLL}_\textrm{gen}=-\mathbb{E}_{{Y_{r}}\sim{P_r}}[\log{P_\theta}(r_{1},\cdots,r_{T})],
\end{align}
where $P_\theta$ is the generated data distribution and $P_r$ is the real data distribution. $\textrm{NLL}_\textrm{oracle}$ is sensitive to the quality, while $\textrm{NLL}_\textrm{div}$ and $\textrm{NLL}_\textrm{gen}$ are sensitive to the diversity.

Since $\textrm{NLL}_\textrm{oracle}$ cannot evaluate the quality of real data, the BLEU scores~\cite{zhu2018texygen} are adopted. For category text generation, the harmonic mean values of the metrics on each category are obtained to evaluate the performance. The repeatable experiment code is made publicly available for further research\footnote{https://github.com/williamSYSU/CatGAN}.

\begin{table}[t]
    \centering
    \caption{The $\textrm{NLL}_\textrm{oracle}$ scores on category text generation. For the $\textrm{NLL}_\textrm{oracle}$ scores, the lower the better.}
\resizebox{0.75\columnwidth}{!}{
\begin{tabular}{|c|c|c|c|}
\hline
Length          & SentiGAN   & CSGAN  & CatGAN\\
\hline
20          & 6.976   & 8.431   & \bf{6.649 $\pm$ 0.097}   \\
40          & 6.821   & 7.621   & \bf{6.498 $\pm$ 0.186}   \\
\hline
    \end{tabular}
}
\label{tb:cat_synthetic}
\end{table}

\begin{table}[!t]
    \centering
    \caption{The performance comparison on MR. $\uparrow$ means higher is better, and $\downarrow$ means lower is better.}
\resizebox{0.75\columnwidth}{!}{
\begin{tabular}{|r|c|c|c|}
\hline
\multicolumn{1}{|c|}{Method}          & SentiGAN   & CSGAN   & CatGAN\\
\hline
BLEU-2 $\uparrow$         & 0.532   & 0.452   & \bf{0.589 $\pm$ 0.041}  \\
BLEU-3 $\uparrow$         & 0.285   & 0.204   & \bf{0.335 $\pm$ 0.032}  \\
BLEU-4 $\uparrow$         & 0.167   & 0.112   & \bf{0.194 $\pm$ 0.028}  \\
BLEU-5 $\uparrow$         & 0.143   & 0.082  & \bf{0.144 $\pm$ 0.028}  \\
\hline
$\textrm{NLL}_\textrm{gen}$ \ $\downarrow$          & 2.436   & 2.912   & \bf{1.619 $\pm$ 0.169}   \\
$\textrm{NLL}_\textrm{div}$ \ $\uparrow$        & 0.484   & 0.254   & \bf{0.535 $\pm$ 0.045}   \\
\hline
    \end{tabular}
}
\label{tb:cat_mr}
\end{table}

\begin{table}[!t]
    \centering
    \caption{The performance comparison on AR.}
\resizebox{0.75\columnwidth}{!}{
\begin{tabular}{|r|c|c|c|}
\hline
\multicolumn{1}{|c|}{Method}          & SentiGAN   & CSGAN   & CatGAN\\
\hline
BLEU-2  $\uparrow$        & 0.870   & 0.879   & \bf{0.987 $\pm$ 0.002}  \\
BLEU-3  $\uparrow$        & 0.801   & 0.674   & \bf{0.943 $\pm$ 0.006}  \\
BLEU-4   $\uparrow$       & 0.691   & 0.442   & \bf{0.867 $\pm$ 0.016}  \\
BLEU-5  $\uparrow$        & 0.554   & 0.256   & \bf{0.751 $\pm$ 0.029}  \\
\hline
$\textrm{NLL}_\textrm{gen}$   $\downarrow$        & 3.374   & 3.197   & \bf{3.104 $\pm$ 0.203}   \\
$\textrm{NLL}_\textrm{div}$   $\uparrow$    & 0.892   & 1.264   & \bf{1.539 $\pm$ 0.050}     \\
\hline
    \end{tabular}
}
\label{tb:cat_ar}
\end{table}
\subsubsection{Datasets.}
Both synthetic and real data are employed to test CatGAN, as in previous works~\cite{guo2018long}.
For category text generation, synthetic data include 20,000 samples, and each 10,000 samples are obtained from different oracle-LSTM~\cite{yu2017seqgan}, and real data include movie reviews (MR)~\cite{socher2013recursive} and amazon reviews (AR)~\cite{mcauley2015image}. MR has two sentiment classes (negative and positive), and AR includes two types of product reviews (book and application).
For general text generation, synthetic data include 10,000 training samples generated by an oracle-LSTM, and real data contain EMNLP2017 WMT News (EN).
All real data employ the same preprocessing as in LeakGAN~\cite{guo2018long}. MR has 4,503 samples, including 3,152 samples for training and 1,351 samples for testing. For AR, each category review includes 100,000 samples for training and 10,000 samples for testing, and each sample may have multiple sentences. EN contains 200,000 training samples and 10,000 test samples.

\subsubsection{Compared Models.}
Several state-of-the-art methods are set as baselines in the experiments.
For category text generation, SentiGAN~\cite{wang2018sentigan} and CSGAN~\cite{li2018generative} are compared with CatGAN $(k=2)$.
For general text generation, four models are compared with CatGAN $(k=1)$, including SeqGAN~\cite{yu2017seqgan}, RankGAN~\cite{lin2017adversarial}, LeakGAN~\cite{guo2018long}, and RelGAN~\cite{nie2018relgan}. The standard MLE training is used for all models before the adversarial training.
For the models which need the temperature, the exponential function $f_{\tau_\text{tar}}(n)=\tau_\text{tar}^{n/N}$ is adopted to increase the temperature, and $\tau_{\text{tar}}$ is set to 1 on synthetic data and 100 on real data.
Adam~\cite{kingma2014adam} is employed to optimize our model.
CatGAN is run with 6 random seeds on all experiments, and the final scores are presented with means and standard deviations (see the \mbox{\emph{Appendix A}} for more detailed settings).

\subsection{Category Text Generation Experiments}
\subsubsection{Synthetic Data Experiments.}
The synthetic data experiments are set with sequence length 20 and 40. $\textrm{NLL}_\textrm{oracle}$ is used to measure the sample quality, and the ground-truth scores are 5.748 and 4.015 for different sequence length, respectively.
In Table~\ref{tb:cat_synthetic}, multiple generators help SentiGAN to obtain competitive results, and CatGAN outperforms SentiGAN by 0.327 and 0.323 on $\textrm{NLL}_\textrm{oracle}$, which illustrates that our model can obtain better quality on all categories.

\begin{table}[t]
    \centering
    \caption{The impact of $\lambda$ on AR.}
\resizebox{0.80\columnwidth}{!}{
\begin{tabular}{|r|c|c|c|c|c|}
\hline
\multicolumn{1}{|c|}{Method}          & 0   & 0.001   & 0.01  & 0.1 & 1\\
\hline
BLEU-2  $\uparrow$    & 0.982 & \textbf{0.987} & 0.982 & 0.980 & 0.982 \\
BLEU-3  $\uparrow$    & 0.942 & \textbf{0.943} & 0.936 & 0.927 & 0.917 \\
BLEU-4  $\uparrow$    & \textbf{0.868} & 0.867 & 0.841 & 0.825 & 0.803 \\
BLEU-5   $\uparrow$   & \textbf{0.757} & 0.751 & 0.712 & 0.682 & 0.654 \\
\hline
$\textrm{NLL}_\textrm{gen}$  $\downarrow$     & 3.577 & 3.104 & 2.902 & 2.552 & \textbf{2.404} \\
$\textrm{NLL}_\textrm{div}$   $\uparrow$    & 1.470 & 1.539 & 1.605 & 1.655 & \textbf{1.689} \\
\hline
    \end{tabular}
}
\label{tb:lambda_ar}
\end{table}

\begin{table}[t]
    \centering
    \caption{The ablation study on AR.}
\resizebox{.95\columnwidth}{!}{
\begin{tabular}{|r|c|c|c|c|}
\hline
\multicolumn{1}{|c|}{Method}   & CatGAN w/o H    & CatGAN w/o T   & CatGAN w/o O     & CatGAN \\
\hline
BLEU-2 $\uparrow$  & 0.977   & 0.986 & 0.979 & \textbf{0.987} \\
BLEU-3 $\uparrow$   & 0.900  & 0.934 & 0.911 & \textbf{0.943} \\
BLEU-4  $\uparrow$  & 0.772  & 0.836 & 0.792 & \textbf{0.867} \\
BLEU-5  $\uparrow$  & 0.613  & 0.703 & 0.638 & \textbf{0.751} \\
\hline
$\textrm{NLL}_\textrm{gen}$ $\downarrow$   & 3.440   & 3.135 & 3.166 & \textbf{3.104} \\
$\textrm{NLL}_\textrm{div}$ $\uparrow$   & 1.524   & 1.555 & \textbf{1.618} & 1.539 \\
\hline
    \end{tabular}
}
\label{tb:ablation_ar}
\end{table}

\subsubsection{Real Data Experiments.}
The real data experiments are conducted on MR and AR. After the same preprocessing, MR consists of 6,216 unique words with the maximum sentence length 15, and AR contains 6,416 unique words with the maximum sentence length 40. The results over generated samples are shown in Table~\ref{tb:cat_mr} and Table~\ref{tb:cat_ar}. On MR, since SentiGAN is designed to generate sentiment text, it shows better results than CSGAN on the BLEU scores. On AR, the sufficient training samples improve the performance of all methods. CSGAN heavily relies on the auxiliary classifier and the RL algorithm, and it shows a significant quality degradation than CatGAN on BLEU while generating long sentences on AR. Compared with the baselines, CatGAN is not limited by the category type of data and obtains the better BLEU scores on both MR and AR, which also shows that CatGAN can catch the dependencies in short and long sentences.
Besides, on AR, CatGAN gets 3.104 on $\textrm{NLL}_\textrm{gen}$ and 1.539 on $\textrm{NLL}_\textrm{div}$, which illustrates that our model can maintain good diversity while significantly improving quality. Actually, optimizing $\textrm{NLL}_\textrm{div}$ based score function can lead to a better $\textrm{NLL}_\textrm{div}$, yet CatGAN still consistently bests $\textrm{NLL}_\textrm{div}$ and an existing metric, $\textrm{NLL}_\textrm{gen}$.

The impact of $\lambda$ is investigated on AR. In the right-hand side of Eq.~\ref{eq:f_obj}, the first term used to measure the quality lies in $[0,1]$, while the second term $\text{NLL}_\text{div}$ usually lies in $[0, 10]$. To balance the sample quality and diversity, $\lambda$ is set to increase from 0 to 1. Table~\ref{tb:lambda_ar} shows that the increase of $\lambda$ triggers the increase of diversity but the degradation of quality, especially for BLEU-4 and BLEU-5. In practice, $\lambda$ is set to 0.001 for CatGAN on all experiments, since it shows a good trade-off between quality and diversity.

\begin{table}[t]
    \centering
    \caption{The $\textrm{NLL}_\textrm{oracle}$ scores on general text generation.}
\resizebox{.95\columnwidth}{!}{
\begin{tabular}{|c|c|c|c|c|c|}
\hline
Length        & SeqGAN   & RankGAN    & LeakGAN   & RelGAN  & CatGAN \\
\hline
20          & 8.736   & 8.247   & 7.038   & 6.680  & \bf{6.377 $\pm$ 0.116} \\
40          & 10.310   & 9.958   & 7.191   & 6.756  & \bf{6.235 $\pm$ 0.131} \\
\hline
    \end{tabular}
}
\label{tb:gene_synthetic}
\end{table}

\begin{figure}[t]
\centering
\includegraphics[width=0.95\columnwidth]{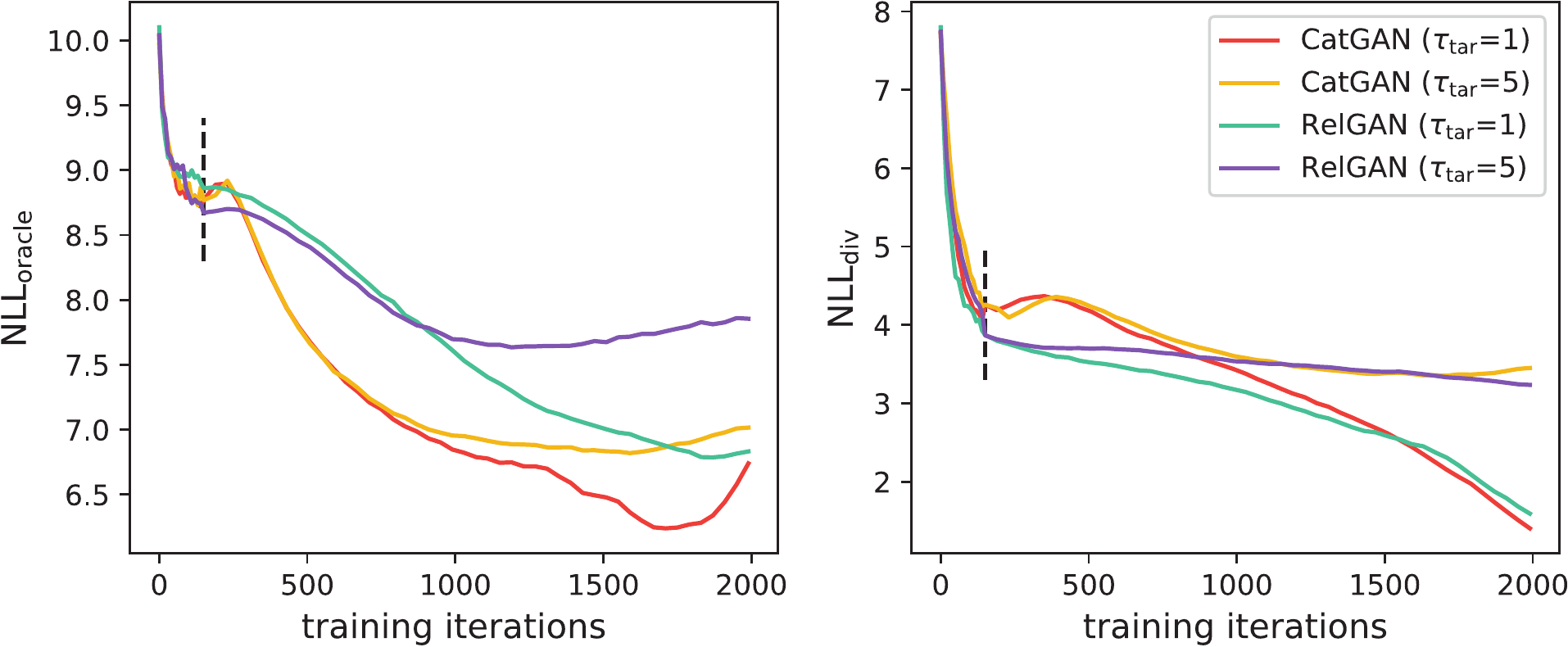}
    \caption{The training curves of CatGAN and RelGAN on the synthetic data of length 20 under different temperatures $\tau_\text{tar} \in \{1, 5\}$. The performances are evaluated with $\text{NLL}_\text{oracle}$ (left) and $\text{NLL}_\text{div}$ (right). The black dotted line represents the end of pre-training.}
    \label{fig:temp}
\end{figure}

\begin{table}[!t]
    \centering
    \caption{The performance comparison on EN.}
\resizebox{.95\columnwidth}{!}{
\begin{tabular}{|r|c|c|c|c|c|}
\hline
\multicolumn{1}{|c|}{Method}          & SeqGAN  & RankGAN    & LeakGAN   & RelGAN  & CatGAN \\
\hline
BLEU-2   $\uparrow$       & 0.777   & 0.727   & 0.826   & 0.881  & \bf{0.954 $\pm$ 0.001}  \\
BLEU-3    $\uparrow$      & 0.491   & 0.435   & 0.645   & 0.705  & \bf{0.804 $\pm$ 0.008}  \\
BLEU-4    $\uparrow$      & 0.261   & 0.209   & 0.437   & 0.501  & \bf{0.603 $\pm$ 0.010}  \\
BLEU-5    $\uparrow$      & 0.138   & 0.101   & 0.272   & 0.319  & \bf{0.402 $\pm$ 0.008} \\
\hline
$\textrm{NLL}_\textrm{gen}$ $\downarrow$         & 2.773   & 3.345   & 2.356   & 2.482  & \bf{2.316 $\pm$ 0.138}  \\
$\textrm{NLL}_\textrm{div}$  $\uparrow$     & 1.695   & 1.178   & 1.291   & 1.117  & \bf{1.716 $\pm$ 0.143}  \\
\hline
    \end{tabular}
}
\label{tb:gene_en}
\end{table}

\begin{table*}[!t]
\centering
\caption{Samples from different methods on MR and AR.}
\resizebox{1.95\columnwidth}{!}{
\begin{tabular}{c | p{0.318\linewidth} | p{0.318\linewidth} | p{0.318\linewidth}}
\hline
 Dataset & \multicolumn{1}{c|}{SentiGAN} & \multicolumn{1}{c|}{CSGAN} & \multicolumn{1}{c}{CatGAN} \\
\hline
\multirow{6}{*}{MR}    &  \bf{Negative}: & \bf{Negative}: & \bf{Negative}: \\
   & a tired, talky a hole in the worst movie.  & goes interesting, and the comedy were interesting. \bf{(Wrong category)} & the premise is intriguing but quickly becomes distasteful and creepy. \\
    &  \bf{Positive}: & \bf{Positive}: & \bf{Positive}: \\
    &   a touching, and politically potent piece of work, a film. &  it 's a treat. \bf{(Short)}&  one of the greatest family-oriented, fantasy-adventure movies ever. \\
\hline
\multirow{8.5}{*}{AR} &  \bf{Book}: & \bf{Book}: & \bf{Book}: \\
&  i have read as the series. i could recommend it to my other walker series. \bf{(Short)} &   this book had an hard time for what's a nice fast read. good character is great reading. what works worth the money. & this is a really good book in a series. the characters are great and they are so easy to read. it is a good read, can't wait for the next book.\\
    & \bf{Application}: & \bf{Application}: & \bf{Application}: \\
        &   i love it. i love it. i would recommend a great game. \bf{(Short)} &   i got this game from amazon it is that i have even it said weather tries.\ \ \ \ \ \ \ \ \ \ \ \ \bf{(Unreadable)} &  great game. i play it until i get to level 3. it 's a nice game for the whole family. my kids too, and it does a great job.\\  
\hline
\end{tabular}
}
\label{tb:case_study}
\end{table*}

For illustrating the effectiveness of TMS and OMS, the ablation study is conducted on AR. The whole hierarchical evolutionary learning algorithm is removed as CatGAN w/o H, TMS is removed from CatGAN as CatGAN w/o T, and OMS is replaced with $l_{G_{\theta}}^{\text{CatRa}}$ to form CatGAN w/o O. The results are shown in Table~\ref{tb:ablation_ar}. Compared with CatGAN, CatGAN w/o T shows the degradation on all BLEU scores and only increases $\textrm{NLL}_\textrm{div}$ by 0.016, which means the worse quality and the similar diversity, respectively. Although CatGAN w/o O achieves competitive sample diversity over CatGAN, it shows a significant degradation on BLEU, which means OMS can effectively guide our model. CatGAN w/o H gets the worse sample quality than CatGAN w/o O, but it still outperforms SentiGAN. The ablation study illustrates that combining TMS and OMS facilitates generating diversified and high-quality samples on real data.

\subsection{General Text Generation Experiments}
The experiments on general text generation are further to show the contribution of the hierarchical evolutionary learning algorithm. General text generation can be considered as the special case of category text generation when $k=1$.
\begin{figure}[!b]
\centering
\includegraphics[width=.95\columnwidth]{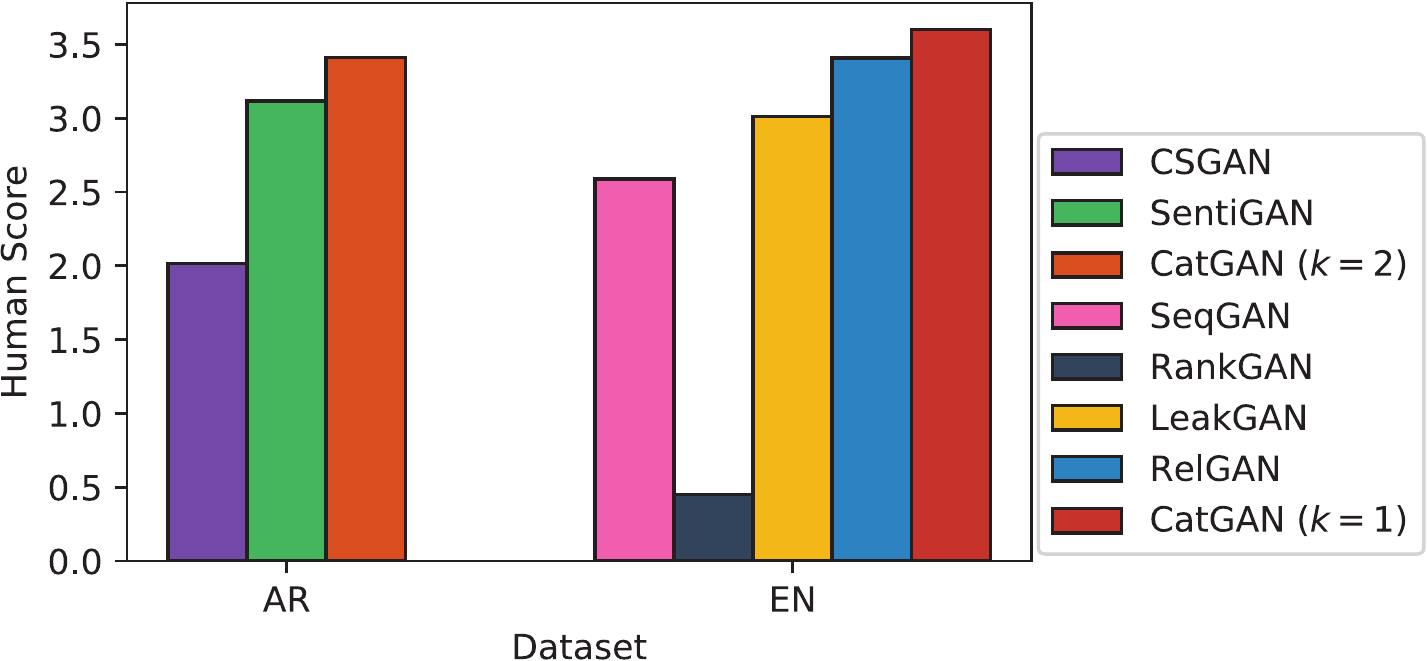}
    \caption{The performance comparison of the human score.}
    \label{fig:human}
\end{figure}
\subsubsection{Synthetic Data Experiments.}
The synthetic data experiments run with sequence length 20 and 40, and the ground-truth $\textrm{NLL}_\textrm{oracle}$ scores are 5.750 and 4.071, respectively. The results are presented in Table~\ref{tb:gene_synthetic}. Compared with all baselines, CatGAN outperforms the best of them by 0.303 and 0.521 on $\textrm{NLL}_\textrm{oracle}$ with different sequence length, respectively, which verifies the better sample quality. Specially, with sequence length 40, CatGAN significantly improves the metric by 0.956 and 3.723 over LeakGAN and RankGAN, respectively, and it illustrates that our model is more powerful to catch long-term dependencies. With the help of the hierarchical evolutionary learning algorithm, CatGAN outperforms RelGAN which also employs the temperature and the Gumbel-Softmax relaxation.

The trade-off between quality and diversity under different temperatures is shown in Fig.~\ref{fig:temp}.
It illustrates that the improvement of quality is always accompanied by the decline in diversity, and higher $\tau_\text{tar}$ brings more diversity.
Compared with RelGAN, CatGAN shows better quality under the same diversity.
With the help of TMS, CatGAN greatly improves the quality over RelGAN under the temperature $\tau_\text{tar}=5$.
The results validate that the hierarchical evolutionary learning algorithm can reduce the impact of mode collapse.

\subsubsection{Real Data Experiments.}
The real data experiments are conducted on the EN dataset. After the preprocessing, EN contains 5,255 unique words with the maximum sentence length 51. Table~\ref{tb:gene_en} shows that CatGAN consistently outperforms other methods on BLEU, which illustrates its power to generate high-quality long sentences. Under the premise of achieving better BLEU scores, our model improves $\textrm{NLL}_\textrm{gen}$ and $\textrm{NLL}_\textrm{div}$ by 0.04 and 0.021 than the best performance of the baselines, which shows CatGAN can maintain the higher sample quality and diversity simultaneously.

\subsection{Human Evaluation}
The human evaluation is conducted for further evaluating the sample quality of generated sentences on AR and EN. The sample quality is measured based on grammatical and semantic correctness, and the detailed protocol is provided in the \mbox{\emph{Appendix D}}. For category text generation, each model randomly generates 100 samples for each category, then these samples with category information are rated by five graduate students with the score from 1 to 5, where 1 means the worst quality and 5 means the best. The harmonic mean values of the average score on each category are shown in Fig.~\ref{fig:human}. For general text generation, the average score over 100 generated sentences from each model is reported. The human evaluation results demonstrate that CatGAN can generate better quality samples than other baselines.

\subsection{Case Study} 
With trained on MR and AR, the generated sentences from SentiGAN, CSGAN, and CatGAN are listed in Table~\ref{tb:case_study}. As shown by the examples, CSGAN shows some problems, such as short length, wrong category and unreadable sentence. Especially on AR, CSGAN lacks the ability for catching the long-term dependencies and generates many unreadable sentences. SentiGAN is capable of obtaining sentiment category text, but it also cannot generate the high-quality long sentences on AR, which may due to the gap between the distributions of two products from AR is larger than the gap of various sentiments. To summarize, CatGAN produces the samples which are longer, more readable and accurate on different categories (see the \mbox{\emph{Appendix E}} for more samples).

\section{Conclusion}
This paper proposes CatGAN for category text generation. 
In order to guide the category-aware model to obtain category samples accurately,
the informative updating signal is provided by measuring the relativistic relation between generated samples and the corresponding real samples on each category.
Besides, a hierarchical evolutionary learning algorithm is developed to train CatGAN and improve generation performance. It allows the model to preserve the well-performing offspring, where the generated category samples can retain diversified and high-quality after each training iteration.
Experimental results on several datasets demonstrate that
CatGAN achieves a better performance than most of the existing state-of-the-art methods on both category text generation and general text generation.

\section{Acknowledgments}
This work is supported by the National Key R\&D Program of China (2018AAA0101203), and the National Natural Science Foundation of China (61673403, U1611262).

\bibliographystyle{aaai}
\fontsize{9.0pt}{10.0pt} \selectfont
\bibliography{reference}

\begin{thebibliography}{}

\bibitem[\protect\citeauthoryear{Bengio \bgroup et al\mbox.\egroup
  }{2015}]{bengio2015scheduled}
Bengio, S.; Vinyals, O.; Jaitly, N.; and Shazeer, N.
\newblock 2015.
\newblock Scheduled sampling for sequence prediction with recurrent neural
  networks.
\newblock In {\em NIPS},  1171--1179.

\bibitem[\protect\citeauthoryear{Caccia \bgroup et al\mbox.\egroup
  }{2018}]{caccia2018language}
Caccia, M.; Caccia, L.; Fedus, W.; Larochelle, H.; Pineau, J.; and Charlin, L.
\newblock 2018.
\newblock Language gans falling short.
\newblock {\em arXiv preprint arXiv:1811.02549}.

\bibitem[\protect\citeauthoryear{Chen \bgroup et al\mbox.\egroup
  }{2018}]{chen2018adversarial}
Chen, L.; Dai, S.; Tao, C.; Zhang, H.; Gan, Z.; Shen, D.; Zhang, Y.; Wang, G.;
  Zhang, R.; and Carin, L.
\newblock 2018.
\newblock Adversarial text generation via feature-mover's distance.
\newblock In {\em NIPS},  4666--4677.

\bibitem[\protect\citeauthoryear{Fedus, Goodfellow, and
  Dai}{2018}]{fedus2018maskgan}
Fedus, W.; Goodfellow, I.; and Dai, A.~M.
\newblock 2018.
\newblock Mask{GAN}: Better text generation via filling in the \_.
\newblock In {\em ICLR}.

\bibitem[\protect\citeauthoryear{Goodfellow \bgroup et al\mbox.\egroup
  }{2014}]{goodfellow2014generative}
Goodfellow, I.; Pouget-Abadie, J.; Mirza, M.; Xu, B.; Warde-Farley, D.; Ozair,
  S.; Courville, A.; and Bengio, Y.
\newblock 2014.
\newblock Generative adversarial nets.
\newblock In {\em NIPS},  2672--2680.

\bibitem[\protect\citeauthoryear{Graves}{2013}]{graves2013generating}
Graves, A.
\newblock 2013.
\newblock Generating sequences with recurrent neural networks.
\newblock {\em arXiv preprint arXiv:1308.0850}.

\bibitem[\protect\citeauthoryear{Gu, Im, and Li}{2018}]{gu2018neural}
Gu, J.; Im, D.~J.; and Li, V.~O.
\newblock 2018.
\newblock Neural machine translation with gumbel-greedy decoding.
\newblock In {\em AAAI}.

\bibitem[\protect\citeauthoryear{Guo \bgroup et al\mbox.\egroup
  }{2018}]{guo2018long}
Guo, J.; Lu, S.; Cai, H.; Zhang, W.; Yu, Y.; and Wang, J.
\newblock 2018.
\newblock Long text generation via adversarial training with leaked
  information.
\newblock In {\em AAAI}.

\bibitem[\protect\citeauthoryear{Hochreiter and
  Schmidhuber}{1997}]{hochreiter1997long}
Hochreiter, S., and Schmidhuber, J.
\newblock 1997.
\newblock Long short-term memory.
\newblock {\em Neural computation} 9(8):1735--1780.

\bibitem[\protect\citeauthoryear{Husz{\'a}r}{2015}]{huszar2015not}
Husz{\'a}r, F.
\newblock 2015.
\newblock How (not) to train your generative model: Scheduled sampling,
  likelihood, adversary?
\newblock {\em arXiv preprint arXiv:1511.05101}.

\bibitem[\protect\citeauthoryear{Jang, Gu, and
  Poole}{2017}]{jang2016categorical}
Jang, E.; Gu, S.; and Poole, B.
\newblock 2017.
\newblock Categorical reparameterization with gumbel-softmax.
\newblock In {\em ICLR}.

\bibitem[\protect\citeauthoryear{Jolicoeur-Martineau}{2018}]{jolicoeur2018relativistic}
Jolicoeur-Martineau, A.
\newblock 2018.
\newblock The relativistic discriminator: a key element missing from standard
  gan.
\newblock {\em arXiv preprint arXiv:1807.00734}.

\bibitem[\protect\citeauthoryear{Kim}{2014}]{kim2014convolutional}
Kim, Y.
\newblock 2014.
\newblock Convolutional neural networks for sentence classification.
\newblock {\em arXiv preprint arXiv:1408.5882}.

\bibitem[\protect\citeauthoryear{Kingma and Ba}{2014}]{kingma2014adam}
Kingma, D.~P., and Ba, J.
\newblock 2014.
\newblock Adam: A method for stochastic optimization.
\newblock {\em arXiv preprint arXiv:1412.6980}.

\bibitem[\protect\citeauthoryear{Li \bgroup et al\mbox.\egroup
  }{2017}]{li2017adversarial}
Li, J.; Monroe, W.; Shi, T.; Jean, S.; Ritter, A.; and Jurafsky, D.
\newblock 2017.
\newblock Adversarial learning for neural dialogue generation.
\newblock {\em arXiv preprint arXiv:1701.06547}.

\bibitem[\protect\citeauthoryear{Li \bgroup et al\mbox.\egroup
  }{2018}]{li2018generative}
Li, Y.; Pan, Q.; Wang, S.; Yang, T.; and Cambria, E.
\newblock 2018.
\newblock A generative model for category text generation.
\newblock {\em Information Sciences} 450:301--315.

\bibitem[\protect\citeauthoryear{Lin \bgroup et al\mbox.\egroup
  }{2017}]{lin2017adversarial}
Lin, K.; Li, D.; He, X.; Zhang, Z.; and Sun, M.-T.
\newblock 2017.
\newblock Adversarial ranking for language generation.
\newblock In {\em NIPS},  3155--3165.

\bibitem[\protect\citeauthoryear{Maddison, Mnih, and
  Teh}{2017}]{maddison2016concrete}
Maddison, C.~J.; Mnih, A.; and Teh, Y.~W.
\newblock 2017.
\newblock The concrete distribution: A continuous relaxation of discrete random
  variables.
\newblock In {\em ICLR}.

\bibitem[\protect\citeauthoryear{McAuley \bgroup et al\mbox.\egroup
  }{2015}]{mcauley2015image}
McAuley, J.; Targett, C.; Shi, Q.; and Van Den~Hengel, A.
\newblock 2015.
\newblock Image-based recommendations on styles and substitutes.
\newblock In {\em SIGIR},  43--52.

\bibitem[\protect\citeauthoryear{Nie, Narodytska, and
  Patel}{2019}]{nie2018relgan}
Nie, W.; Narodytska, N.; and Patel, A.
\newblock 2019.
\newblock Rel{GAN}: Relational generative adversarial networks for text
  generation.
\newblock In {\em ICLR}.

\bibitem[\protect\citeauthoryear{Santoro \bgroup et al\mbox.\egroup
  }{2018}]{santoro2018relational}
Santoro, A.; Faulkner, R.; Raposo, D.; Rae, J.; Chrzanowski, M.; Weber, T.;
  Wierstra, D.; Vinyals, O.; Pascanu, R.; and Lillicrap, T.
\newblock 2018.
\newblock Relational recurrent neural networks.
\newblock In {\em NIPS},  7299--7310.

\bibitem[\protect\citeauthoryear{Semeniuta, Severyn, and
  Gelly}{2018}]{semeniuta2018accurate}
Semeniuta, S.; Severyn, A.; and Gelly, S.
\newblock 2018.
\newblock On accurate evaluation of gans for language generation.
\newblock {\em arXiv preprint arXiv:1806.04936}.

\bibitem[\protect\citeauthoryear{Socher \bgroup et al\mbox.\egroup
  }{2013}]{socher2013recursive}
Socher, R.; Perelygin, A.; Wu, J.; Chuang, J.; Manning, C.~D.; Ng, A.; and
  Potts, C.
\newblock 2013.
\newblock Recursive deep models for semantic compositionality over a sentiment
  treebank.
\newblock In {\em EMNLP},  1631--1642.

\bibitem[\protect\citeauthoryear{Tucker \bgroup et al\mbox.\egroup
  }{2017}]{tucker2017rebar}
Tucker, G.; Mnih, A.; Maddison, C.~J.; Lawson, J.; and Sohl-Dickstein, J.
\newblock 2017.
\newblock Rebar: Low-variance, unbiased gradient estimates for discrete latent
  variable models.
\newblock In {\em NIPS},  2627--2636.

\bibitem[\protect\citeauthoryear{Vaswani \bgroup et al\mbox.\egroup
  }{2017}]{vaswani2017attention}
Vaswani, A.; Shazeer, N.; Parmar, N.; Uszkoreit, J.; Jones, L.; Gomez, A.~N.;
  Kaiser, {\L}.; and Polosukhin, I.
\newblock 2017.
\newblock Attention is all you need.
\newblock In {\em NIPS},  5998--6008.

\bibitem[\protect\citeauthoryear{Wang and Wan}{2018}]{wang2018sentigan}
Wang, K., and Wan, X.
\newblock 2018.
\newblock Senti{GAN}: Generating sentimental texts via mixture adversarial
  networks.
\newblock In {\em IJCAI},  4446--4452.

\bibitem[\protect\citeauthoryear{Wang \bgroup et al\mbox.\egroup
  }{2019}]{wang2019evolutionary}
Wang, C.; Xu, C.; Yao, X.; and Tao, D.
\newblock 2019.
\newblock Evolutionary generative adversarial networks.
\newblock {\em IEEE Transactions on Evolutionary Computation}.

\bibitem[\protect\citeauthoryear{Williams}{1992}]{williams1992simple}
Williams, R.~J.
\newblock 1992.
\newblock Simple statistical gradient-following algorithms for connectionist
  reinforcement learning.
\newblock {\em Machine learning} 8(3-4):229--256.

\bibitem[\protect\citeauthoryear{Yu \bgroup et al\mbox.\egroup
  }{2017}]{yu2017seqgan}
Yu, L.; Zhang, W.; Wang, J.; and Yu, Y.
\newblock 2017.
\newblock Seq{GAN}: Sequence generative adversarial nets with policy gradient.
\newblock In {\em AAAI}.

\bibitem[\protect\citeauthoryear{Zhang \bgroup et al\mbox.\egroup
  }{2017}]{zhang2017adversarial}
Zhang, Y.; Gan, Z.; Fan, K.; Chen, Z.; Henao, R.; Shen, D.; and Carin, L.
\newblock 2017.
\newblock Adversarial feature matching for text generation.
\newblock In {\em ICML},  4006--4015.

\bibitem[\protect\citeauthoryear{Zhu \bgroup et al\mbox.\egroup
  }{2018}]{zhu2018texygen}
Zhu, Y.; Lu, S.; Zheng, L.; Guo, J.; Zhang, W.; Wang, J.; and Yu, Y.
\newblock 2018.
\newblock Texygen: A benchmarking platform for text generation models.
\newblock In {\em SIGIR},  1097--1100.

\end{thebibliography}

\newpage
\ \ \ 


\section*{Supplementary material}

# Appendix A

**Experimental Settings**

- In the CNN based discriminator, for the sake of simplicity, filter windows of sizes are set to be $\{2, 3, 4, 5\}$, and the dimension of each feature map is set to be 300.
- For the RMC based generator, we set the number of heads to be 2, memory slots to be 1 and memory size to be 512.
- Following the settings of the baselines, the batch size is set to be 64 during the whole training procedure. The embedding size of the input token for the generator is set to be 32 and that for the discriminator to be 64.
- For the pre-training phase, the same as all baseline methods, the standard MLE training is used to train the generator with the learning rate of 1e-2 for 150 epochs.
- For the adversarial training, the learning rate is empirically set to be 1e-4 for both generator and discriminator, and the maximum number of iterations is set to be $N = 5000$.
- In the previous work (Wang et al. 2019), the number of parents is set to be 1 or 4. In our experiments, with increasing the number of parents, the training time and the trainable parameters will multiply, while the performance does not have a significant improvement. Moreover, there have been already $|\mathbb{T}| \times |\mathbb{O}| = 6$ mutation directions, and the possible solutions can be well explored in the parameter space. Thus, the number of parents is simply set to be 1 in our model, and one best-performing generator is selected in each evolution.

# Appendix B

## Pseudo Code

---

**Algorithm 1** The algorithm of CatGAN.

---

**Require:** The generator $G_\theta$; the discriminator $D_\phi$; the number of categories $k$; the discriminator's updating steps per iteration $n_D$; the mutation direction $\mathcal{M}^{(\tau,\omega)}$ with the temperature $\tau$ and the objective $\omega$.*

1: Initialize $G_\theta$ and $D_\phi$ with random weights $\theta$ and $\phi$
2: Pre-train $G_\theta$ on $k$ categories of samples at the same time
3: **for** $n = 1, \cdots, N$ **do**
4:    **for** $\mathcal{M}^{(\tau,\omega)}$ in $\mathbb{T} \times \mathbb{O}$ **do**                                                               ▷ Variation and Evaluation
5:       Sample a batch of $Y_\theta^c \sim P_\theta^c$ (generated data), and a batch of $Y_r^c \sim P_r^c$ (real data) on each category $c = \{1, \cdots, k\}$
6:       $G_\theta$ produces offspring $G_\theta^{(\tau,\omega)}$ via mutation $\mathcal{M}^{(\tau,\omega)}$, that is, updating generator $G_\theta$ parameters via Eq. 4 or Eq. 5
7:       Evaluate the offspring $G_\theta^{(\tau,\omega)}$ with $\mathcal{F}_{\text{temp}}^{(\tau,\omega)}$ and $\mathcal{F}_{\text{obj}}^{(\tau,\omega)}$
8:    **end for**
9:    **for** $\omega$ in $\mathbb{O}$ **do**                                                                                                      ▷ Selection in Stage$_{\text{temp}}$
10:       $\{\mathcal{F}_{\text{temp}}^{(\tau_\omega^1,\omega)}, \cdots, \mathcal{F}_{\text{temp}}^{(\tau_\omega^{|\mathbb{T}|},\omega)}\} \leftarrow \text{sort}(\{\mathcal{F}_{\text{temp}}^{(\tau,\omega)}\})$
11:       Preserve the best-performing offspring $G_\theta^{(\tau_\omega^1,\omega)}$
12:    **end for**
13:    $\{\mathcal{F}_{\text{obj}}^{(\tau_{\omega_1}^1,\omega_1)}, \cdots, \mathcal{F}_{\text{obj}}^{(\tau_{\omega_{|\mathbb{O}|}}^1,\omega_{|\mathbb{O}|})}\} \leftarrow \text{sort}(\{\mathcal{F}_{\text{obj}}^{(\tau_\omega^1,\omega)}\})$                          ▷ Selection in Stage$_{\text{obj}}$
14:    Preserve the best-performing offspring $G_\theta^{(\tau_{\omega_1}^1,\omega_1)}$ as the new parent, that is $G_\theta \leftarrow G_\theta^{(\tau_{\omega_1}^1,\omega_1)}$
15:    **for** $d = 1, \cdots, n_D$ **do**
16:       Sample a batch of $Y_\theta^c \sim P_\theta^c$ (generated data), and a batch of $Y_r^c \sim P_r^c$ (real data) on each category $c = \{1, \cdots, k\}$
17:       Update discriminator $D_\phi$ parameters via Eq. 1, Eq. 2 and Eq. 3
18:    **end for**
19: **end for**

*Default values $k = 2, n_D = 5, \mathbb{T} = \{f_{\tau_{\text{tar}}}(n-1), f_{\tau_{\text{tar}}}(n), f_{\tau_{\text{tar}}}(n+1)\}, \mathbb{O} = \{l_{G_\theta}^{\text{CatRS}}, l_{G_\theta}^{\text{CatRa}}\}, \tau \in \mathbb{T}, \omega \in \mathbb{O}$.

## Objective of Discriminator

$$l_{D_\phi}^{\text{CatRa}} = \sum_{c=1}^{k} \mathcal{L}^{\text{Ra}}(Y_r^c, Y_\theta^c) + \mathcal{L}^{\text{Ra}}(Y_r^{\text{all}}, Y_\theta^{\text{all}}), \tag{1}$$

$$\mathcal{L}^{\text{Ra}}(Y_r, Y_\theta) = -\mathbb{E}_{Y_r \sim P_r}\left[\log\left(\bar{D}_\phi(Y_r)\right)\right] - \mathbb{E}_{Y_\theta \sim P_\theta}\left[\log\left(1 - \bar{D}_\phi(Y_\theta)\right)\right], \tag{2}$$

$$\bar{D}_\phi(Y) = \begin{cases} \text{sigmoid}(D_\phi(Y) - \mathbb{E}_{Y_\theta \sim P_\theta}[D_\phi(Y_\theta)]) & \text{if } Y \text{ is real} \\ \text{sigmoid}(D_\phi(Y) - \mathbb{E}_{Y_r \sim P_r}[D_\phi(Y_r)]) & \text{otherwise.} \end{cases} \tag{3}$$

## Objective of Generator

$$l_{G_\theta}^{\text{CatRa}} = -l_{D_\phi}^{\text{CatRa}}, \tag{4}$$

$$l_{G_\theta}^{\text{CatRS}} = -\sum_{c=1}^{k} \mathbb{E}_{\substack{Y_r^c \sim P_r^c \\ Y_\theta^c \sim P_\theta^c}}[\log(\text{sigmoid}(D_\phi(Y_\theta^c) - D_\phi(Y_r^c)))] - \mathbb{E}_{\substack{Y_r^{\text{all}} \sim P_r^{\text{all}} \\ Y_\theta^{\text{all}} \sim P_\theta^{\text{all}}}}[\log(\text{sigmoid}(D_\phi(Y_\theta^{\text{all}}) - D_\phi(Y_r^{\text{all}})))]. \tag{5}$$

# Appendix C

**Evolutionary Temperature**

The comparison of temperature curves between the evolutionary temperature and the monotone increasing temperature (i.e., the exponential temperature obtained by $f_{\tau_{\text{tar}}}(n) = \tau_{\text{tar}}^{n/N}$) is shown in Fig. 1. Although the two strategies of controlling temperature are similar in overall, the evolutionary temperature does not strictly follow the monotone increasing process.

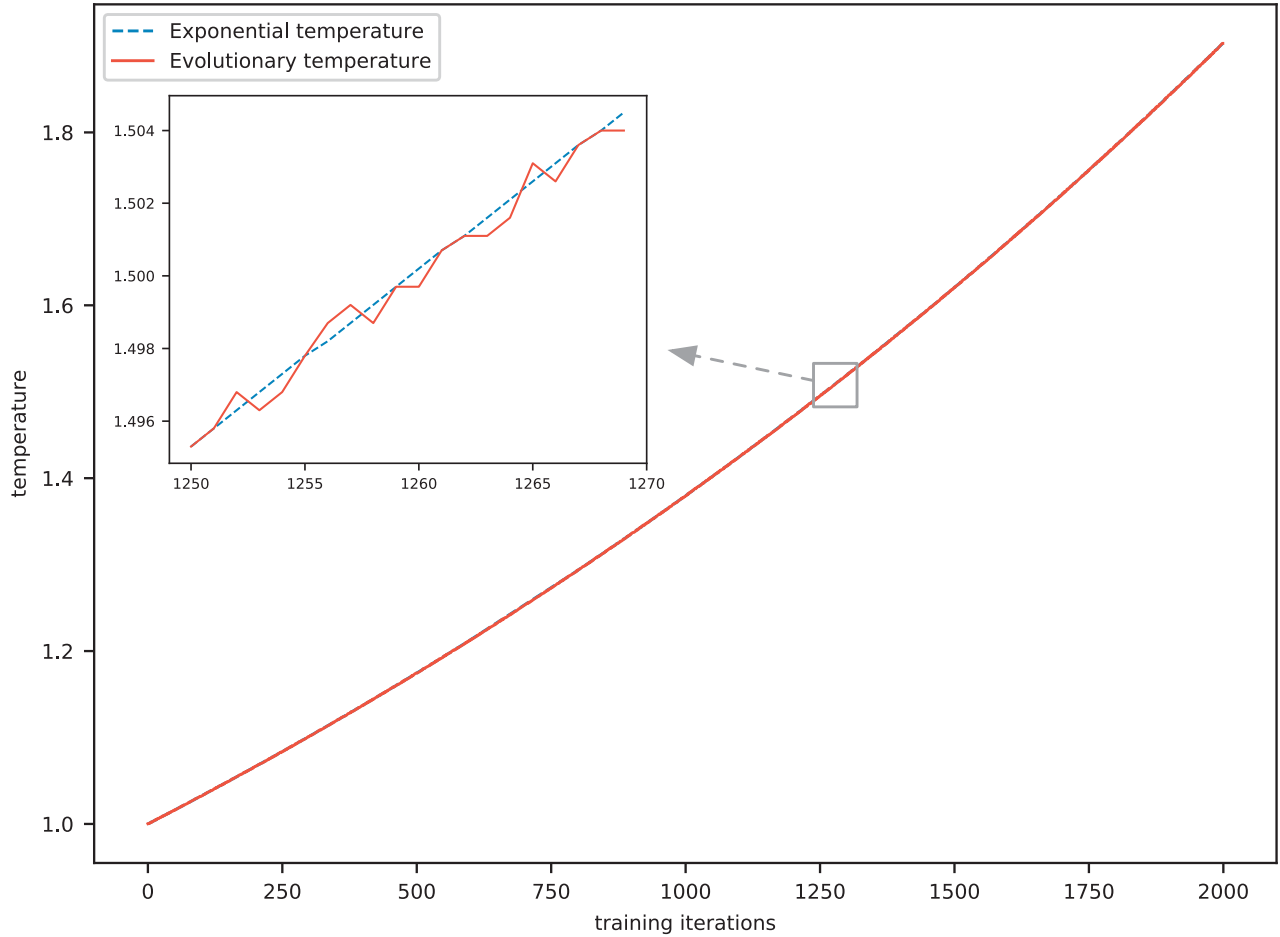

Figure 1: Comparison of the temperature curves when $\tau_{\text{tar}} = 5$ on synthetic data with length 20. The exponential temperature is obtained by $\tau_{\text{tar}}^{n/N}$ for each iteration. The evolutionary temperature is selected by the hierarchical evolutionary learning algorithm. The overall figure shows the temperature curves over 2000 training iterations.

# Appendix D

## Human Evaluation Protocols

The category text quality in human evaluation is based on grammatical and semantic correctness. The detailed protocols, as in existing SentiGAN and RelGAN, are reported in Table 1 and Table 2, where the criterion consists of category correctness, grammatical correctness, and meaningfulness. Each model generates the same number of samples for each category, then these samples with category information are rated by students. The harmonic mean values of the average score on each category are presented in Fig. 3. For general text generation, the category correctness criterion can be removed from the protocol.

Table 1: The human evaluation protocol for category text generation. Supposing the right category is Application on AR.

| Scale | Criterion & Example |
|---|---|
| 5 - Excellent | Right category, grammatically correct, and making sense. For example, *"everyone in my family loves this game. i also love it and always play it to pass the time, because the graphics are so cute."* |
| 4 - Good | Right category, some small grammatical errors, and mostly making sense. For example, *"i downloaded this on my kindle for a long time, and who have to wait for the coins to play the game."* |
| 3 - Fair | Right category, major grammatical errors, but conveying some meanings. For example, *"i hate this app only lets me take another opinion it hard to play without buying stuff."* |
| 2 - Poor | Right category, not making sense, but some locally meaningful parts. For example, *"i get this app and my granddaughter get the stuff, i don't better are a purpose."* |
| 1 - Unacceptable | Wrong category, or a random collection of words. For example, *"i a great way to kill time and who watches the reviews ever, a very easy book to read."* |

Table 2: The human evaluation protocol for general text generation.

| Scale | Criterion & Example |
|---|---|
| 5 - Excellent | Grammatically correct, and making sense. For example, *"if England wins the World Cup next year, it will be the most significant result the sport has seen in more than a decade."* |
| 4 - Good | Some small grammatical errors, and mostly making sense. For example, *"it is useful to have had a doctor who forced her to release him a couple of days before she was cleared."* |
| 3 - Fair | Major grammatical errors, but conveying some meanings. For example, *"even then once again there ' s a sign of that stuff is going on the way to work on christmas eve."* |
| 2 - Poor | Not making sense, but some locally meaningful parts. For example, *"we go to work for the moment in life their eyes and, i have been a different race on to go."* |
| 1 - Unacceptable | A random collection of words. For example, *"i go com com com, i on on on play can go go."* |

# Appendix E

## Generated Samples on Real Data

Table 3: Randomly chosen 13 generated samples trained on Amazon Reviews with $\tau_{\text{tar}} = 100$.

**Book:**
i have the book after reading the first two books and enjoyed it. i found it a good reading. the plot and character development was excellent. i would recommend this book.
this is a great series. i would recommend it to your readers. i enjoyed it. i enjoyed reading it as well i found it interesting after reading.
it is a great book. it is a very good book. it had me surprised. it 's just the best book. i 'm so glad i bought it. i definitely recommend it to anyone.
well i never finished in perfect condition! i am an really fan of the books in this series. so far each has a better plot line.
loved the ending was a great read i ca n't wait to read the next book. i hope to get the next one soon.
the book was really interesting. i think it was a fairly good read. i hope the next book is as good, so to each another book really.
a mystery – kept me on the edge of my seat. lots of twists and turns that i could n't put down. ca n't wait to read the next book.
although, i got a little bored the first couple of the book. the romance is a little to much of a great read! it was a cute story, i did.
i still think i should try this book. it was a book club. i believe good but i do n't think i will i just keep it.
i love the colorful characters and they just keep you reading. it was a great read. looking forward to reading some of this authors books.
again, it took me a long time to read this book. it was a really enjoyable read. i hope this writer is the author i get to care!
this was a great book. the story is fast paced and not too engrossed. excellent book. i recommend it to anyone with an entertaining book.
once i did, i loved this book. i was hooked on this series, i ca n't put it down. you get to know the characters until you are just about the story.

**Application:**
my son downloaded this to my kindle. i have had it for a while, and it 's a fun so far. i like it a lot.
i did n't do it. it kept my attention for a really long time, it is a great game. it might be a lot more interesting. i would recommend this game to anyone.
i have a kindle fire on some of the time. i enjoyed this game. i was so much fun to play. it was a good story.
it worked great but and definitely a huge waste of my memory. i was just really disappointed with this app. hope there is a computer version.
i always thought this would be fun but since it is kind of boring that because after about 3 times i thought about for something else but it changed.
very entertaining game, from the first one to a very good slot game! i really like all i could not put it down.
not one of those was the games i was looking for. the game was very hard. i do not enjoy this game. i like all the different games.
nice product. the music is great. it has another option with a challenge. i thought that it would be fair but not nearly so great.
i really like this game. it is a good time waster and you can do. i will keep it. the levels are great. i 'm glad that i was able to play it.
the game is useful to get my brain working. it is fun, but it is not one of the more interesting games that i have played. it 's a great time killer.
i got it when it was a free app, and it is a great game, it is a great game to play. it 's a very fast game and you can not wait to get to it.
if you like music i would totally recommend this for anyone out there. i gave to my daughter because i was hoping for something similar to her.
this app would be great for kids to play it and it is fun and interesting that you can enjoy the bad guys.

Table 4: Randomly chosen 13 generated samples trained on EMNLP2017 WMT News with $\tau_{\text{tar}} = 100$.

" over the past 10 years, i know how much you have to be here and come into the end of one, " he said in a 2015 interview in the wake.
" to be a police, in these four years they have a difficult time and have to come out here and get better, " he said.
" this means having medical that ' s not right, so if you and your children will be very confident, and they have to be better at big. " he said.
the local police said, according to law, and that was a case of the government ' s health.
" it was a big day, but i think i ' ve got to stay with them and it ' s not where we ' ll be at the right time, " she said.
" you know what they ' ve been looking at across the world, and you don ' t want to do that, " she said.
" i am sure he will be one of them , mr.trump said in a statement to the recent, although the others won ' t build a new team. " he said.
the new york has seen just about 1 . 5 million in the coming days, with more than 20 percent of the population of germany.
the bank said it would agree that the 2016 trade took its day ' s stock was up to 2.5 % in the wake of june after brexit.
" i don ' t think he can be a problem forever over the next two or three weeks, so i didn ' t have to go " he said.
i have to be working at a bad place and try to make this happen, so i ' m very hard on myself.
" i ' ve heard that he wants to be there as an actor but now that i ' m going to hold him for the next game, " he said.
" that ' s why we have the reports before we get the chance to be able to actually reach the world east, " he said.
clinton also holds for $ 1 . 8 million in the last, and on wednesday, though, they have a hard time trying to get the trump and bernie sanders in here.
" amazon also wants to cut down its economic in the next year, when more than 1 , 700 people have been in the us, now has been the worst in a year. " the report said.



\end{document}